\documentclass{article} 
\usepackage[a4paper,headheight=20pt,top=25mm,bottom=35mm,left=35mm,right=25mm]{geometry}

\usepackage{arxiv}
\usepackage[T1]{fontenc}    
\usepackage{fancyhdr}       
\usepackage{hyphenat}
\usepackage{hyperref}
\usepackage{amsfonts}
\usepackage{amsmath}
\usepackage{amsthm}
\usepackage{graphicx}
\usepackage{subfigure}
\usepackage{enumitem}
\usepackage{array}
\usepackage{booktabs}

\usepackage{tikz}

\newcolumntype{C}[1]{>{\centering\let\newline\\\arraybackslash\hspace{0pt}}p{#1}}

\title{\centering Reinforcement Learning with Intrinsic Affinity\\for Personalized Prosperity Management}

\author{Charl Maree%
\thanks{Second affiliation: Chief Technology Office, Sparebank 1 SR-Bank, Stavanger, Norway.} 
\, and Christian W. Omlin \\ 
Center for Artificial Intelligence Research\\ 
University of Agder\\ 
Grimstad, Norway \\ 
\texttt{\{charl.maree,christian.omlin\}@uia.no} \\ }

\begin{document}
\maketitle

\begin{abstract}
The common purpose of applying reinforcement learning (RL) to asset management is the maximization of profit. The extrinsic reward function used to learn an optimal strategy typically does not take into account any other preferences or constraints. We have developed a regularization method that ensures that strategies have global intrinsic affinities, i.e., different personalities may have preferences for certain assets which may change over time. We capitalize on these intrinsic policy affinities to make our RL model inherently interpretable. We demonstrate how RL agents can be trained to orchestrate such individual policies for particular personality profiles and still achieve high returns.
\end{abstract}


\section{Introduction}
Effective customer engagement is a requisite for modern financial service providers that are adopting advanced methods to increase the level of personalization of their services \cite{stefanel19}. Although artificial intelligence (AI) has become a ubiquitous tool in financial technology \cite{fernandes19}, research in the field has yet to significantly advance levels of personalization \cite{maree_clustering}. Asset management is an active research topic in AI for finance; however, the research opportunities presented by the need for personalized services are usually neglected \cite{Millea21}. Whereas personalized investment advice is typically based on questionnaires, we propose the use of micro-segmentation based on spending behavior. 
Traditionally, customer segmentation has been grounded in demographics which provide only a coarse segmentation \cite{Smith56}; it fails to capture nuanced differences between individuals with the potential for undesirable ramifications, e.g. discrimination in credit scoring based on postal code \cite{Barocas16}. Micro-segmentation, however, provides a more sophisticated classification which can improve the quality of banking services \cite{Mousaeirad20, Apeh11}.

We develop a personal asset manager that invests in a portfolio of asset classes according to individual personality profiles as manifested by their spending behavior. The result is a hierarchical system of reinforcement learning (RL) agents in which a high-level agent orchestrates the actions of five low-level agents with global intrinsic affinities for certain asset classes. These affinities derive from prototypical personality traits. For instance, personality traits with a higher affinity for risk may, as a general rule, prefer high-volatility assets. 

Explainability and interpretability are key in sensitive industries, such as finance \cite{arrieta2020}; they form the basis for understanding and trust and have not yet been adequately addressed \cite{matz21, Cao21}. In short, our agents' policies are regularized by predefined prior action distributions which imprint characteristic behaviors, making their policies inherently interpretable on three levels: (1) they use salient features extracted from customer spending behavior, (2) the affinity of the prototypical agents, and (3) their orchestration to achieve personal investment advice. Our contribution is therefore twofold: we demonstrate how RL agents can be made inherently interpretable through their intrinsic affinities, and their application and value to personalized asset management.

\section{Background and Related Work} \label{sec:background}

Spending patterns are a predictor of financial personality \cite{glad19}. In their paper, the authors trained a random forest to predict customer personalities from their classified financial transactions, using a prevalent taxonomy of personality traits: openness, conscientiousness, extraversion, agreeableness, and neuroticism. Although they achieved only a modest predictive accuracy, a subsequent study found that spending patterns over time expose salient information that is obscured in non-temporal form \cite{Tovanich21}; the authors in this study used the same personality model, but added temporal patterns such as variability of the amount, persistence of the category in time, and burstiness---the intermittent changes in frequency of an event. Recurrent neural networks (RNNs) are able to extract this salient information when predicting personality traits from financial transactions \cite{maree_clustering}. In, \cite{maree_understanding}, we gained an understanding of these extracted features by interpreting the dynamics of the RNN state space through a set of attractors. Understanding model behavior is crucial in industries such as personal finance \cite{matz21}. In their study, \cite{matz21} extracted rules from three classes of models---linear regression, logistic regression, and random forests---which not only exposed the spending patterns most indicative of personality traits, but also aided in model improvement.  

In RL, agents learn to solve problems by tentation; they maximize the expected rewards resulting from their actions in an environment \cite{Sutton98}. Environments can have complex dynamics that result in sophisticated policies that are opaque to their developers, who have no influence over what these agents learn \cite{heuillet2021a, garcia2015}. Intrinsic motivation enables agents to learn behaviors that are detached from the expected rewards of the environment \cite{aubret19}. It is a strategy that was developed to address the challenge of exploration in environments with sparse rewards \cite{andres22}. One such approach is Kullback-Leibler (KL) policy regularization in which the objective function is regularized by the KL-divergence between the current policy and a predefined prior \cite{galashov2018}. Policy regularization has been shown to be helpful and never detrimental to convergence \cite{Vieillard2020}. Although most policy regularization methods aim to improve learning performance, they can also control the learning process and imbue the policy with an intrinsic behavior \cite{maree_yourway}. Here, the objective function is regularized with a predefined prior action distribution that defines a desirable characteristic:
\begin{align} \label{eqn:obj_reg}
    J(\theta) &= \mathbb{E}_{s,a \sim \mathcal{D}} \left[ R(s,a) \right] - \lambda L \\ \nonumber%
    L &= \frac{1}{M} \sum_{j=0}^{M} \Big[ \mathbb{E}_{a \sim \pi_\theta} \left[ a_j \right] - (a_{j} \vert \pi_0(a)) \Big]^2
\end{align}
$J(\theta)$ is the learning objective as a function of the model parameters $\theta$, $R(s,a)$ is the expected reward for state $s$ and action $a$ as sampled from a replay buffer $\mathcal{D}$, and $\lambda$ is a scaling hyperparameter for the regularization term $L$, which is the mean square difference between the observed action distribution and the action distribution given a regularization prior $\pi_0$. The efficacy of this approach was demonstrated by instilling an inherent characteristic behavior in agents that navigate a grid. These agents learned to either prefer left turns, right turns, or to avoid going straight by taking a zig-zag approach to their destination. In contrast to constrained RL, which \emph{avoids} certain states, the policy regularization in \cite{maree_yourway} \emph{encourages} certain actions irrespective of the state and is a new direction for RL.

Hierarchical reinforcement learning (HRL) decomposes problems into low-level subtasks that are learned by relatively simple agents for the purpose of either improved performance or explainability \cite{Pateria22, Levy19}. Larger problems are solved by choreographing these subtasks through an orchestration agent that learns the high-level dynamics of its environment \cite{Hengst2010}. HRL has, for instance, been used to control a robotic arm: while low-level agents learned simple tasks such as moving forward / backward or picking up / placing down, an orchestration agent learned to retrieve objects on a surface by choreographing these tasks \cite{marzari21,beyret19}. The agents were not only efficient at learning, but their policies were more easily interpreted by human experts. In \cite{Kulkarni16}, the authors used HRL to train a hierarchical set of agents to play a game. Their low-level agents learned to solve simple tasks such as ``pick up a key'' or ``open a door'' while receiving extrinsic rewards from the environment. A high-level agent then orchestrated these sub-tasks and received intrinsic rewards generated by a critic based on whether or not larger objectives were met.


\section{Methodology} \label{sec:methodology}
\begin{figure}[!ht]
    \centering
    \includegraphics[width=0.75\linewidth]{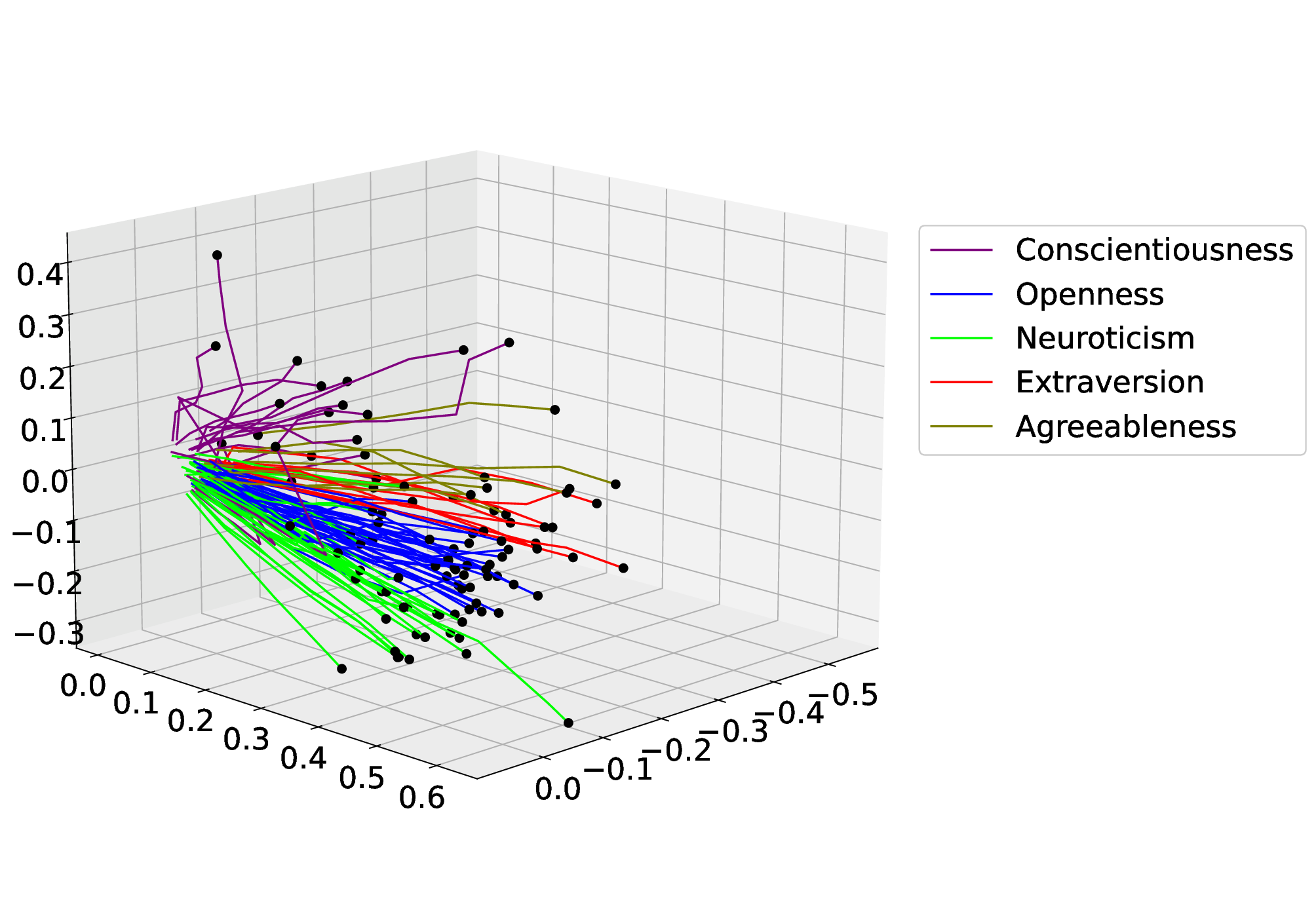}
    \caption{Clustering behavior of trajectories in the state space of a RNN. Each trajectory represents a customer's spending behavior in time and is labelled according to the customer's dominant personality trait.}
    \label{fig:micro-seg_features}
\end{figure}

\begin{figure}[!ht]
    \centering
    \includegraphics[width=0.8\linewidth]{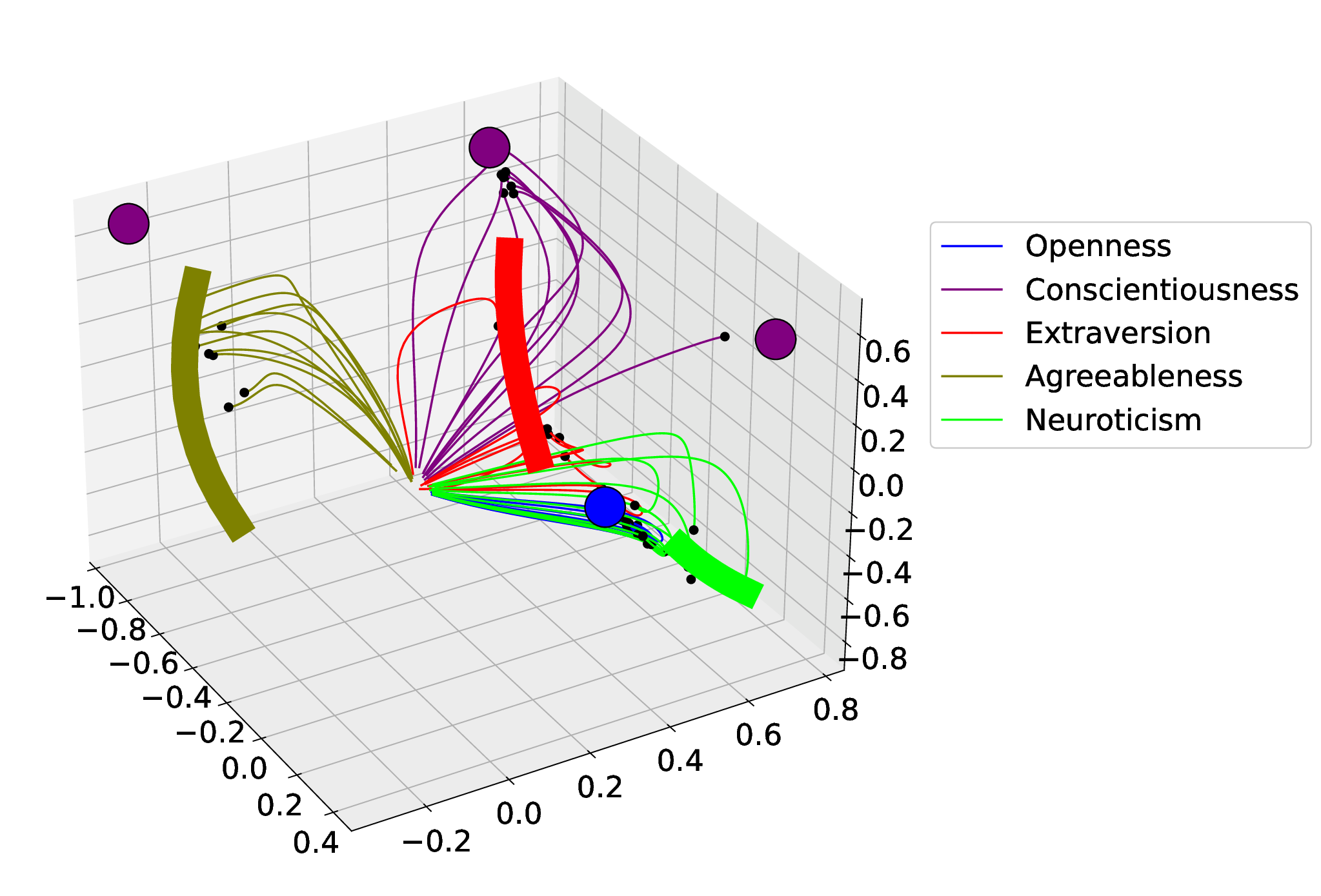}
    \caption{The locations of a set of attractors in the state space of a RNN. There are point and line attractors that are labelled according to the customers' corresponding dominant personality traits. The shown trajectories are the resulting state space features when the first time-step of a subset of customers' spending histories are copied to create 100 time steps; this represents a spending history that is constant in time and allows the trajectories to asymptotically converge on to their respective attractors. }
    \label{fig:attractors}
\end{figure}

We have previously developed a three-node RNN that predicts customer personalities from an input vector of their classified financial transactions \cite{maree_clustering}. This input vector consists of six annual time steps, each consisting of 97 transaction classes; the values in each time step add up to one and are the fraction of a customer's annual spending per transaction category. The RNN output is a five-dimensional personality vector; its values are the degrees of membership in each of five personality traits: openness, conscientiousness, extraversion, agreeableness, and neuroticism. We use the feature trajectories from this model's state space---shown in Figure~\ref{fig:micro-seg_features}---to represent a customer's spending behavior over time.
Each behavioral trajectory represents an individual customer and is labeled according to their most dominant personality trait: the trait with the greatest value in the personality vector. These trajectories form clusters in the state space, which form hierarchical sub-clusters along the successive levels of dominance of the personality traits. This hierarchical clustering provides a means of micro-segmenting customers according to their spending behavior in time. We then explained these behavioral trajectories by reproducing them using a linear regression model, and we interpreted them through locating a number of attractors that govern the dynamics of the state space \cite{maree_understanding}. We located these attractors by mapping the RNN output space into the state space through inverse regression. Using this mapping, and the maximum \emph{reachable} values in the output space, based on the known range of the dimensions in the state space ($[-1,1]$), we extrapolated the final locations (attractors) of the behavioral trajectories. Formally:
\begin{alignat*}{2} \label{eqn:mapped_dim}
&\mathcal{O} &&= \mathcal{D} \cdot \omega_{inv} - \bf{0}^P \cdot \omega_{inv} \\
&\mathcal{D} &&= diag \left\{ \max_{1 \leq i \leq |K|}{O_{i,j}, \ j \in [1..P]} \right\} \\
&\omega_{inv} &&= (O^TO)^{-1} \cdot (O^TS) \\
&O &&= S \cdot \omega_{out}
\end{alignat*}
where $\mathcal{O}$ is the projection of the output space into the state space, $\bf{0}^P$ is the zero vector or origin of the output space, $O$ is the reachable output space, $S = [-1,1]^3$ is the reachable state space, $\omega_{out}$ is the vector of weights of the RNN's output layer, $P=5$ is the number of output dimensions, and $K$ is the number of points used to map the output hypercube. We corroborated these theoretical locations with the observed destinations of the trajectories; for each customer, we repeated the first time step 100 times, thereby creating trajectories that asymptotically converge to their attractors. We thus determined that trajectories converge towards the attractor associated with their most dominant personality trait. If a customer's spending behavior changes such that a different personality trait becomes dominant, their trajectory changes direction accordingly towards the new appropriate attractor. Figure~\ref{fig:attractors} shows these attractors in the RNN state space, with the extended spending trajectories, converging towards the appropriate attractors. 
There are three point attractors for the personality trait conscientiousness, towards which trajectories converge depending on their initial conditions. Agreeableness, extraversion, and neuroticism each have a single line attractor, while trajectories that classified as openness converge towards a single point attractor. Each basin of attraction forms a cluster of trajectories, which each form a hierarchy of sub-clusters along successive levels of dominance of personality traits. This is the interpretation of the trajectory dynamics.

\begin{figure}[!ht]
    \centering
    \includegraphics[width=0.98\linewidth]{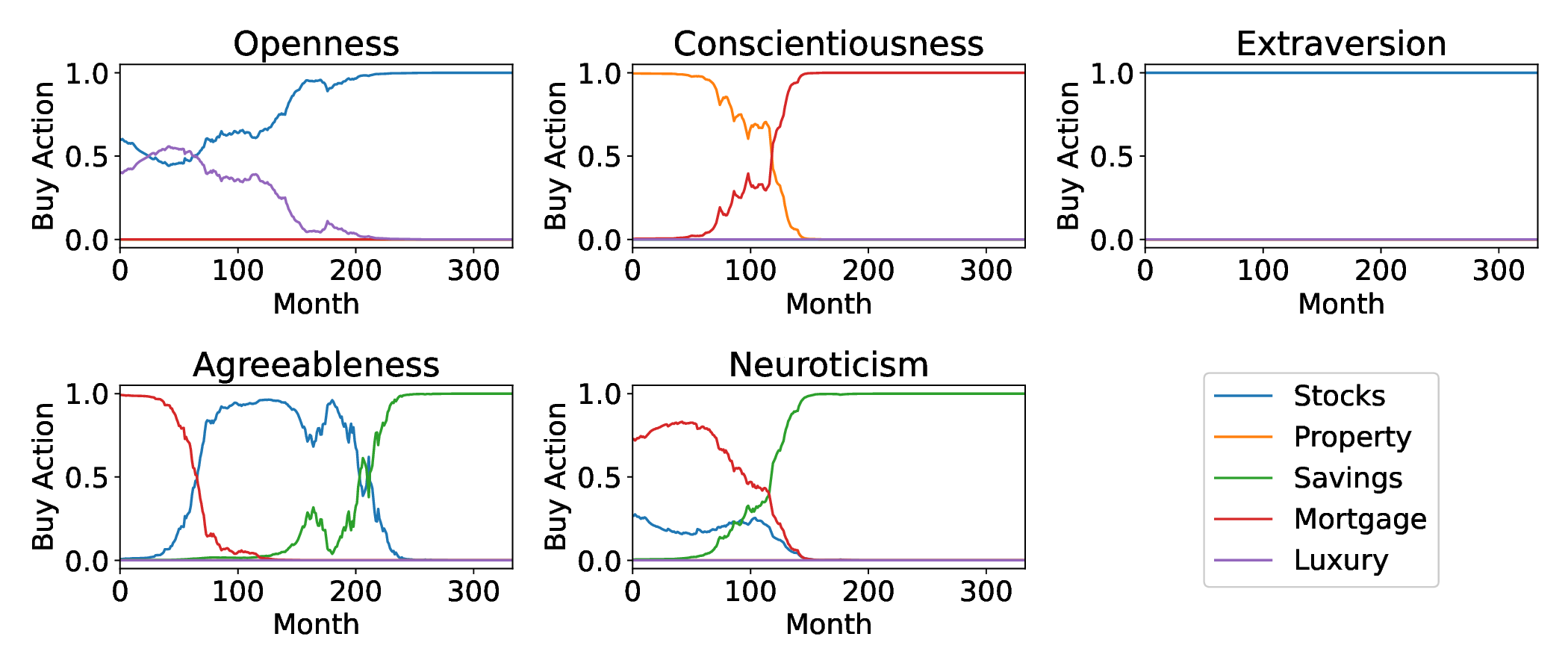}
    \caption{Action distributions of the five prototypical agents over a 30 year time period: between the ages of 30 and 60. Each figure represents the investment actions taken by one of the prototypical agents, who each associates with a single personality trait. Each line represents the fractional investment into a labelled class of assets across the time period, e.g. the conscientiousness agent initially invests solely in property and subsequently pays off mortgage debt, while the extraversion agent always invests the entire monthly amount in stocks.}
    \label{fig:prototypical_actions}
\end{figure}

Interpretable RL can be used for investment that matches personality \cite{maree_can}. In this preliminary study, we had trained multiple RL agents to invest monthly contributions in different financial assets: stocks, property, savings accounts, mortgage payments, and luxury expenses. We obtained assets prices from the S\&P~500 index \cite{yf_sp500}, the Norwegian property index \cite{ssb_prop}, and the Norwegian interest rate index \cite{nb_int}. With the help of a panel of experts from a major bank, we ranked these asset classes according to the following properties: expected returns, typical liquidity, capital prerequisite, typical risk, and novelty. We then associated the personality traits with these properties, as shown in Table~\ref{tab:tab_asset_class_prop}. 
\begin{table}[ht]
    \centering
    \caption{A set of asset class properties and their associations with the five personality traits: openness, conscientiousness, extraversion, agreeableness, and neuroticism. The values are in the set $\left\{ n \in \mathbb{Z} \ \vert -2 \leq n \leq 2 \right\}$ and indicate a strong negative, slightly negative, neutral, slightly positive and strong positive association, respectively.}
    \label{tab:tab_asset_class_prop}
    \begin{tabular}{lccccc}
        \toprule
        Asset class property & Open. & Cons. & Extra. & Agree. & Neur. \\
        \midrule
        Expected returns & 1 & 1 & 2 & 1 & 1 \\
        Liquidity & 2 & -1 & 2 & 1 & 2 \\
        Low capital prerequisite & 0 & -1 & 1 & 1 & 1 \\
        Low Risk & -1 & 2 & -1 & 1 & 2 \\
        Novelty & 2 & 0 & 2 & 0 & -1 \\
        \bottomrule
    \end{tabular}
\end{table}
The result showed that, for instance, the openness trait might value asset liquidity and novelty; because of their openness to new experiences, they might prefer to have cash readily at hand when such an opportunity presents itself, or they might value assets that in themselves contain novelty. Another example is that the conscientiousness trait might prefer assets with low risk. Combining the coefficients in Table~\ref{tab:tab_asset_class_prop} with the asset property rankings, we mapped the association of each personality trait with the asset classes, as shown in Table~\ref{tab:coefficients}.
\begin{table}[!ht]
 \centering
    \caption{Coefficients, in the range $[-1,1]$, associating asset classes to prototypical personality traits: openness, conscientiousness, extraversion, agreeableness, and neuroticism.} \label{tab:coefficients}
    \footnotesize
    \begin{tabular}{lccccc}
        \toprule
        Asset type & Open. & Cons. & Extra. & Agree. & Neuro. \\
        \midrule
        Savings account  & -0.11 &  0.08 & -0.15 &  0.51 &  0.68 \\
        Property funds & -0.15 &  0.32 & -0.22 & -0.36 & -0.24 \\
        Stock portfolio   &  0.82 & -0.61 &  0.95 &  0.42 &  0.12 \\
        Luxury expenses   &  0.16 & -0.51 & -0.07 & -0.80 & -0.81 \\
        Mortgage repayments & -0.72 &  0.72 & -0.52 &  0.23 &  0.25 \\
        \bottomrule
    \end{tabular}
\end{table}
These coefficients reveal that, for example, the extraversion trait has a high preference for stocks, while the conscientiousness agent prefers a combination of mortgage repayments and property investment. When scaled so that they add up to one and their minimum values equal to zero, these coefficients were the regularization priors $\pi_0$ in Equation~\ref{eqn:obj_reg}; we regularized the objective functions of five prototypical agents to instill intrinsic affinities for certain asset classes. Each agent learned an investment strategy associated with one of the five personality traits, which is the interpretation of their policies. Figure~\ref{fig:prototypical_actions} shows these strategies, where each agent acted in an environment in which they invested a fixed monthly amount of 10~000 Norwegian Kroner (NOK) for 30 years. These prototypical agents clearly learned unique strategies for investing. The openness agent initially preferred luxury expenditures, in line with their openness to new experiences, and later purely invested in stocks, which had scored high in novelty. In contrast, the conscientiousness agent preferred to reduce risk through property investment, followed by resolute mortgage payments. These are the low-level policies which we intend to orchestrate into personalized asset investment strategies; customers have varying degrees of membership in each of the five personality traits, resulting in unique preferences for assets that may change over time.

Based on the premise that there is a causal relationship between personality-matched spending and happiness \cite{glad16}, we hypothesize that investment matched with personality can similarly increase happiness. Our goal is therefore to learn, through high-level RL orchestration, the optimum strategies that match customers' unique financial personalities. Our RL agent orchestrates the actions of low-level prototypical agents according to customers' extracted behavioral trajectories (Figure~\ref{fig:micro-seg_features}). With actions adding up to one, representing the fraction of the investment amount allocated to each low-level agent, it maximizes the following reward function: the inner product between the values of asset holdings and the customer's preference for each asset type. This preference is calculated as the inner product of the customer's personality vector---the set of five values representing their degrees of membership in each of the personality traits---and the set of coefficients relating each asset type with each personality trait (Table~\ref{tab:coefficients}). This reward measures the correlation between spending behavior and investment strategy, which we call the \emph{satisfaction index}; the higher the satisfaction index, the higher the correlation between spending behavior and investment strategy. We then instill an intrinsic RL affinity through a regularization prior---the customer's personality vector scaled by its sum---in a deep deterministic policy gradient algorithm (DDPG - \cite{lillicrap2019}). The actor consisted of three vanilla RNN nodes and an output layer of five actions with a softmax activation. The critic had a similar three-node RNN layer for the states which, concatenated with the actions, were succeeded by a 1000 node feed-forward layer and a single output node with no activation function. We found that three RNN nodes provided consistently high total rewards, which is consistent with findings that RNN architectures generally perform well in low-dimensional representations \cite{mahes19}. We tuned our hyperparameters using a one-at-a-time parameter sweep to reach the following optima: the actor and critic learning rates were 0.005 and 0.01, respectively, the target network update parameter $\tau$ was 0.05, the discount factor $\gamma$ was 0.95, and the regularization scaling factor $\lambda$ was 5.  

In summary, we trained five low-level RL agents to invest in a set of assets according to prototypical personality traits \cite{maree_can}. We now combine their actions using a high-level RL agent that uses a customer's spending behavior as manifested in their transaction history. This spending behavior is evinced in behavioral trajectories that we extracted from the state space of a RNN that predicts personality from financial transactions \cite{maree_clustering}. We illustrate this in a flow diagram in Figure~\ref{fig:flow_diagram}.

\tikzstyle{rect} = [rectangle, rounded corners, 
                    text width=2.0cm, 
                    text centered, draw=black, fill=gray!10]
\tikzstyle{circ} = [circle, font=\tiny,
                    text width=0.5cm,
                    text centered, draw=black]
\usetikzlibrary{positioning, calc}

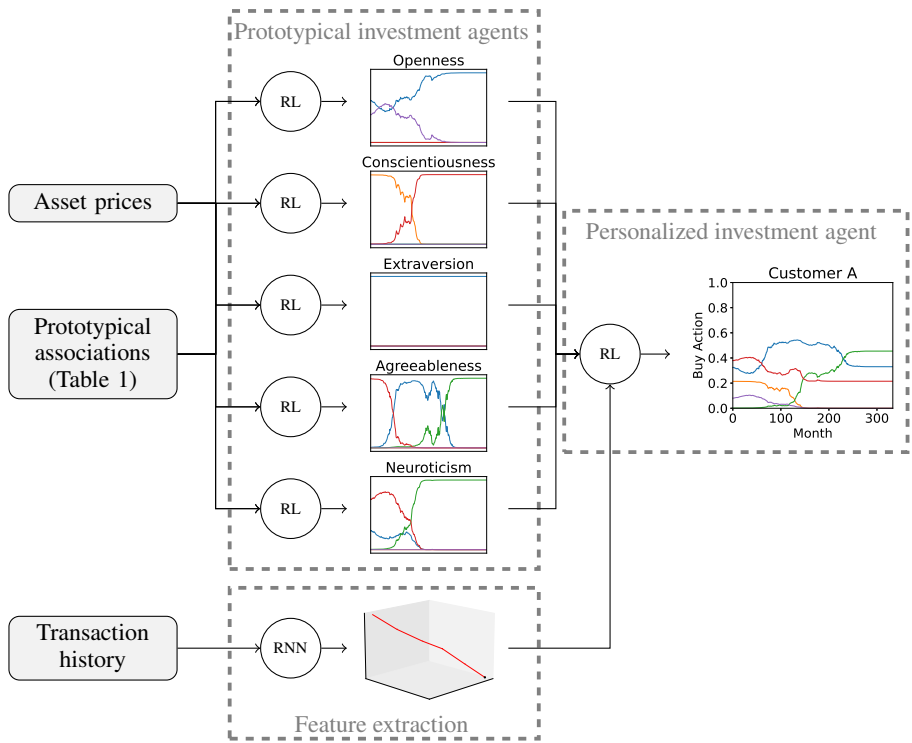
\begin{figure}[!ht]
 \centering
 \footnotesize
 \begin{tikzpicture}
  \node (assets) [rect] {Asset prices};
  \node (associ) [rect, below of=assets, yshift=-1cm] {Prototypical associations\\(Table~\ref{tab:tab_asset_class_prop})};
  \node (trans) [rect, below of=assets, yshift=-4.9cm] {Transaction history};
  
  \node (rlo) [circ, right of=assets, xshift=1.6cm, yshift=1.35cm] {RL};
  \node (rlc) [circ, right of=assets, xshift=1.6cm, yshift=0cm] {RL};
  \node (rle) [circ, right of=assets, xshift=1.6cm, yshift=-1.35cm] {RL};
  \node (rla) [circ, right of=assets, xshift=1.6cm, yshift=-2.7cm] {RL};
  \node (rln) [circ, right of=assets, xshift=1.6cm, yshift=-4.05cm] {RL};
  \node (rlf) [circ, right of=associ, xshift=5.8cm, yshift=0cm] {RL};
  \node (rnn) [circ, right of=trans, xshift=1.6cm, yshift=0cm] {RNN};
  
  \node (open) [right of=rlo, xshift=0.8cm] {\includegraphics[width=1.9cm]{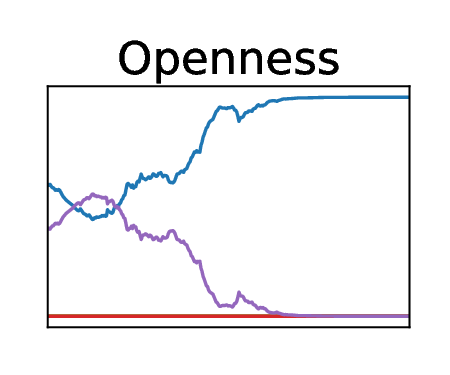}};
  \node (cons) [right of=rlc, xshift=0.8cm] {\includegraphics[width=1.9cm]{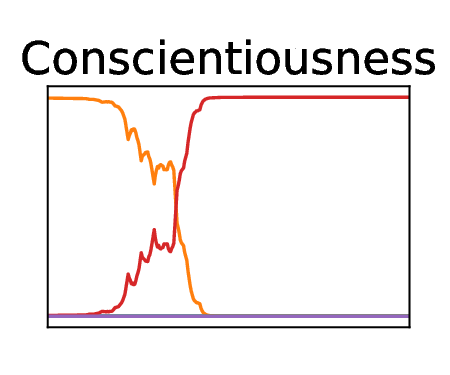}};
  \node (extr) [right of=rle, xshift=0.8cm] {\includegraphics[width=1.9cm]{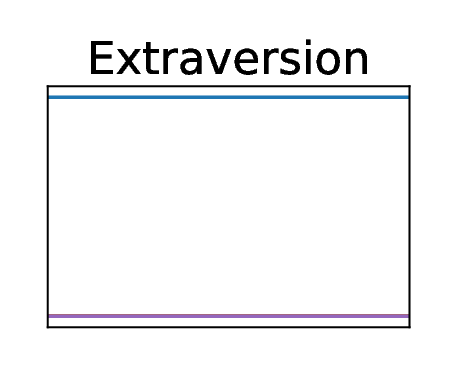}};
  \node (agre) [right of=rla, xshift=0.8cm] {\includegraphics[width=1.9cm]{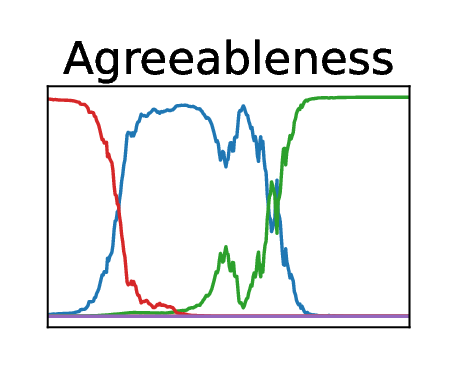}};
  \node (neur) [right of=rln, xshift=0.8cm] {\includegraphics[width=1.9cm]{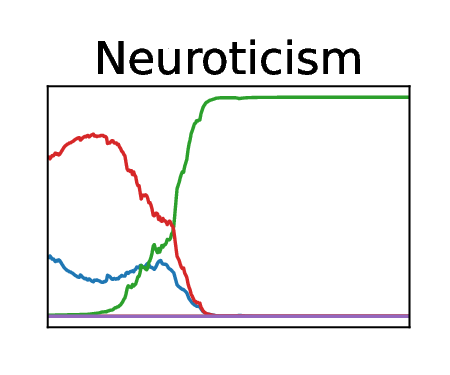}};
  
  \node (feat) [right of=rnn, xshift=0.8cm] {\includegraphics[width=1.9cm]{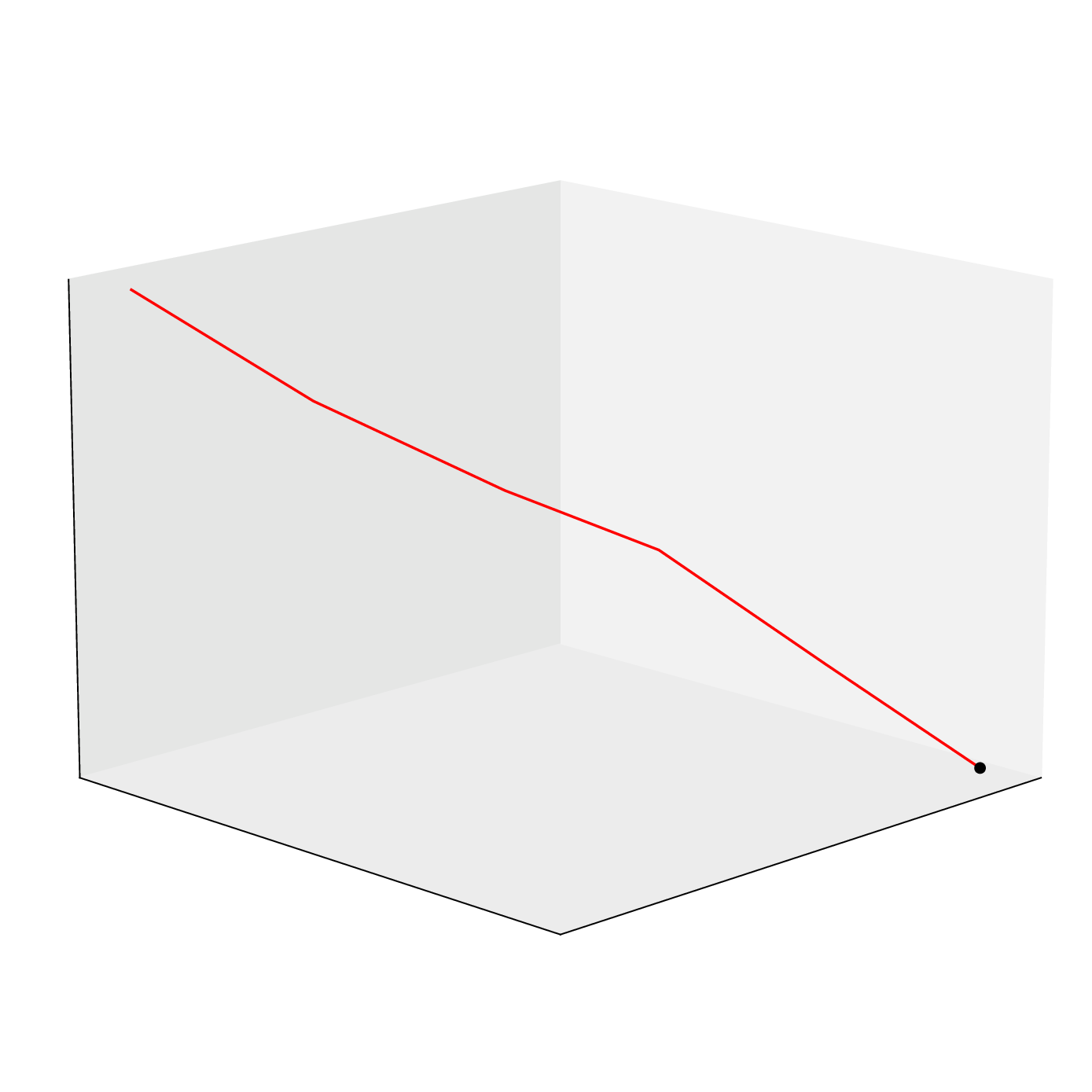}};
  
  \node (orch) [right of=associ, xshift=8.2cm, yshift=0cm] {\includegraphics[width=3.0cm]{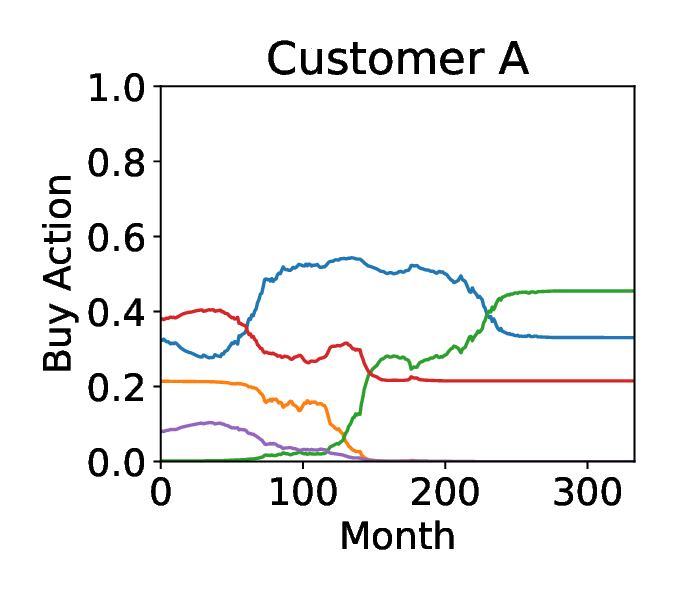}};
  
  \draw[dashed,gray,line width=0.5mm] ($(rlo.west) + (-4mm,12mm)$) rectangle ($(neur.east) + (4mm,-8mm)$);
  \node[gray] at ($(rlo.west) + (16mm,9mm)$) {Prototypical investment agents};
  
  \draw[dashed,gray,line width=0.5mm] ($(rlf.west) + (-2mm,19mm)$) rectangle ($(orch.east) + (-1mm,-13mm)$);
  \node[gray] at ($(rlf.west) + (20mm,16mm)$) {Personalized investment agent};
  
  \draw[dashed,gray,line width=0.5mm] ($(rnn.west) + (-4mm,8mm)$) rectangle ($(feat.east) + (4mm,-12mm)$);
  \node[gray] at ($(rnn.west) + (16mm,-10mm)$) {Feature extraction};
  
  \draw[->] (assets.east) -| ($(assets.east) + (5mm,0)$) |- (rlo.west) [black];
  \draw[->] (assets.east) -| ($(assets.east) + (5mm,0)$) |- (rlc.west) [black];
  \draw[->] (assets.east) -| ($(assets.east) + (5mm,0)$) |- (rle.west) [black];
  \draw[->] (assets.east) -| ($(assets.east) + (5mm,0)$) |- (rla.west) [black];
  \draw[->] (assets.east) -| ($(assets.east) + (5mm,0)$) |- (rln.west) [black];
  \draw[->] (associ.east) -| ($(assets.east) + (5mm,0)$) |- (rlo.west) [black];
  \draw[->] (associ.east) -| ($(assets.east) + (5mm,0)$) |- (rlc.west) [black];
  \draw[->] (associ.east) -| ($(assets.east) + (5mm,0)$) |- (rle.west) [black];
  \draw[->] (associ.east) -| ($(assets.east) + (5mm,0)$) |- (rla.west) [black];
  \draw[->] (associ.east) -| ($(assets.east) + (5mm,0)$) |- (rln.west) [black];
  \draw[->] (rlo.east) |- (open.west) [black]; 
  \draw[->] (rlc.east) |- (cons.west) [black]; 
  \draw[->] (rle.east) |- (extr.west) [black]; 
  \draw[->] (rla.east) |- (agre.west) [black]; 
  \draw[->] (rln.east) |- (neur.west) [black]; 

  \draw[->] (trans.east) -- (rnn) [black];
  \draw[->] (rnn.east) -- (feat) [black];
  
  \draw[->] (open.east) -| ($(orch.west) - (15mm,0)$) |- (rlf.west) [black];
  \draw[->] (cons.east) -| ($(orch.west) - (15mm,0)$) |- (rlf.west) [black];
  \draw[->] (extr.east) -| ($(orch.west) - (15mm,0)$) |- (rlf.west) [black];
  \draw[->] (agre.east) -| ($(orch.west) - (15mm,0)$) |- (rlf.west) [black];
  \draw[->] (neur.east) -| ($(orch.west) - (15mm,0)$) |- (rlf.west) [black];
  
  \draw[->] (feat.east) -| (rlf.south) [black];
  \draw[->] (rlf.east) |- (orch.west) [black]; 
  
 \end{tikzpicture}
 \caption{Information flow diagram illustrating how our system uses financial transactions to generate personalized investment advice. We use hierarchical RL agents with intrinsic affinity to orchestrate prototypical investment strategies to match personal financial preferences.}
 \label{fig:flow_diagram}
\end{figure}

\section{Results}
\begin{figure}[!ht]
    \centering
    \includegraphics[width=8.0cm]{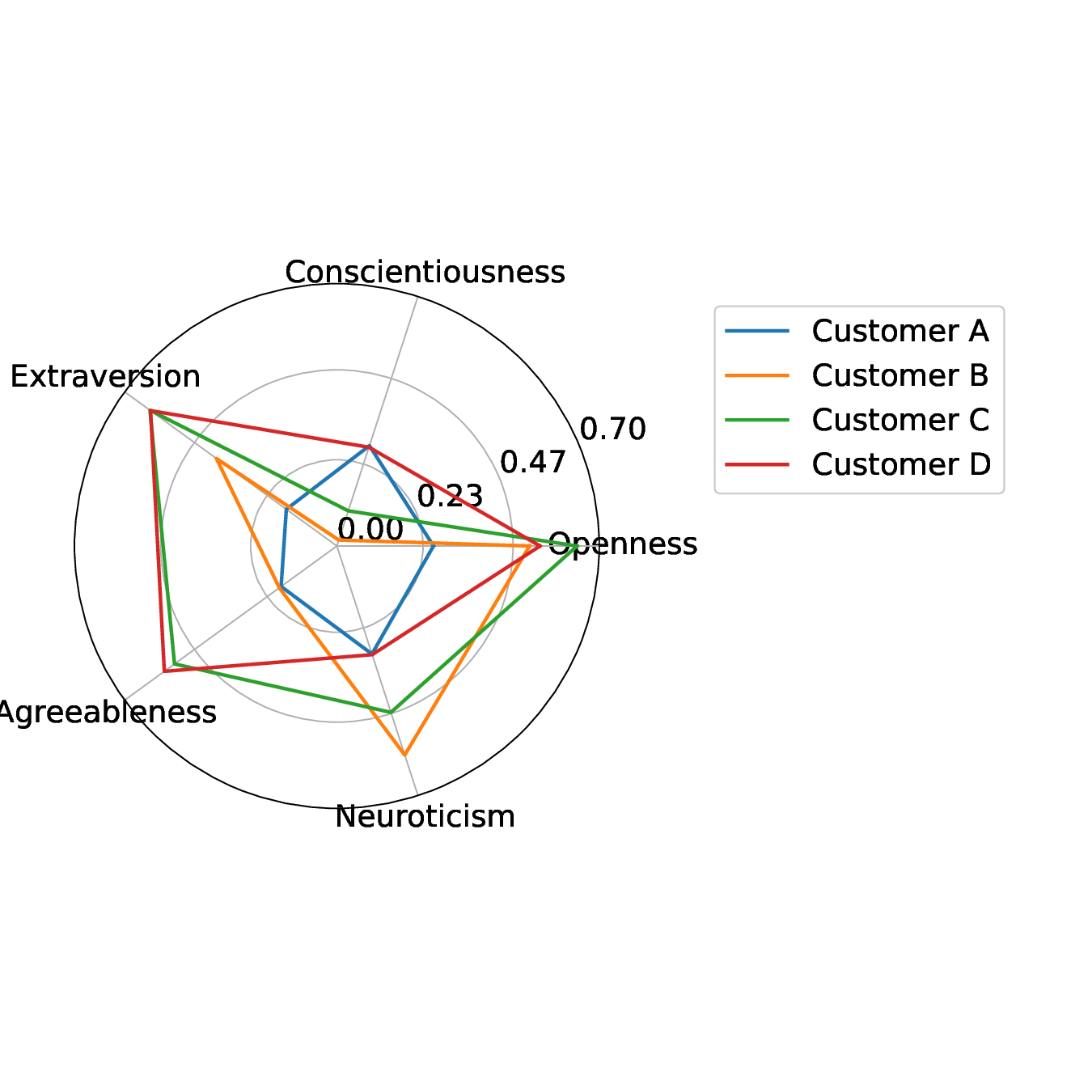}
    \caption{The personality vectors representing the personality traits of four customers. Each colored line represents a customer and each axis on the radar plot represents a personality trait. The values on the axes are in the range $[0,1]$ and represent the customers' degree of membership in each of the personality traits.}
    \label{fig:customers}
\end{figure}

We selected four customers for whom we trained personal orchestration agents; their personality vectors are visualized in Figure~\ref{fig:customers}.  
Customer A has a relatively balanced profile, with low variation in the values of their personality vector, which also has relatively small values. This contrasts with Customer B who scores high in neuroticism and openness, Customer C who scores high in openness and extraversion, and Customer D who scores high in extraversion, agreeableness, and openness. Their respective regularization priors are shown in Table~\ref{tab:reg_priors}. 
\begin{table}[ht]
    \centering
    \caption{Regularization priors used during training of the orchestration agents of four customers, named A through D. Each row represents the regularization prior $\pi_{0,i}$ for one of the orchestration agents $i \in [A,D]$. The values represent the fraction of investment amount allocated to each prototypical low-level agent: openness, conscientiousness, extraversion, agreeableness and neuroticism.}
    \begin{tabular}{cccccc}
         \toprule 
         Prior & Open. & Cons. & Extra. & Agree. & Neur. \\
         \midrule 
         $\pi_{0,A}$ & 0.22 & 0.24 & 0.14 & 0.15 & 0.25 \\
         $\pi_{0,B}$ & 0.30 & 0.01 & 0.23 & 0.11 & 0.35 \\
         $\pi_{0,C}$ & 0.27 & 0.04 & 0.26 & 0.23 & 0.20 \\
         $\pi_{0,D}$ & 0.23 & 0.12 & 0.27 & 0.25 & 0.13 \\
         \bottomrule
    \end{tabular}
    \label{tab:reg_priors}
\end{table}
The regularization prior for Agent A $\pi_{0,A}$ (the agent for Customer A) consequently has a low variation in its values while $\pi_{0,B}$ assigned the highest weight to neuroticism and openness, $\pi_{0,C}$ assigned the highest weight to openness and extraversion, and $\pi_{0,D}$ assigned the highest weight to extraversion, agreeableness, and openness. 

\begin{figure}[!ht]
    \centering
    \subfigure{
        \includegraphics[width=.48\linewidth]{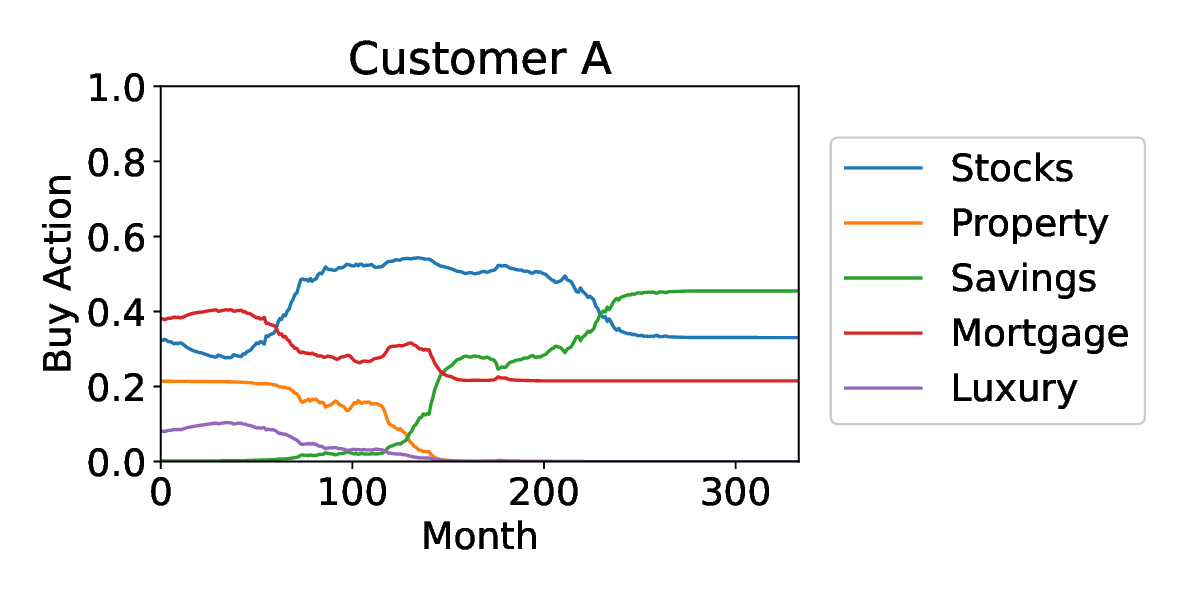}
    }\hfill
    \subfigure{
        \includegraphics[width=.48\linewidth]{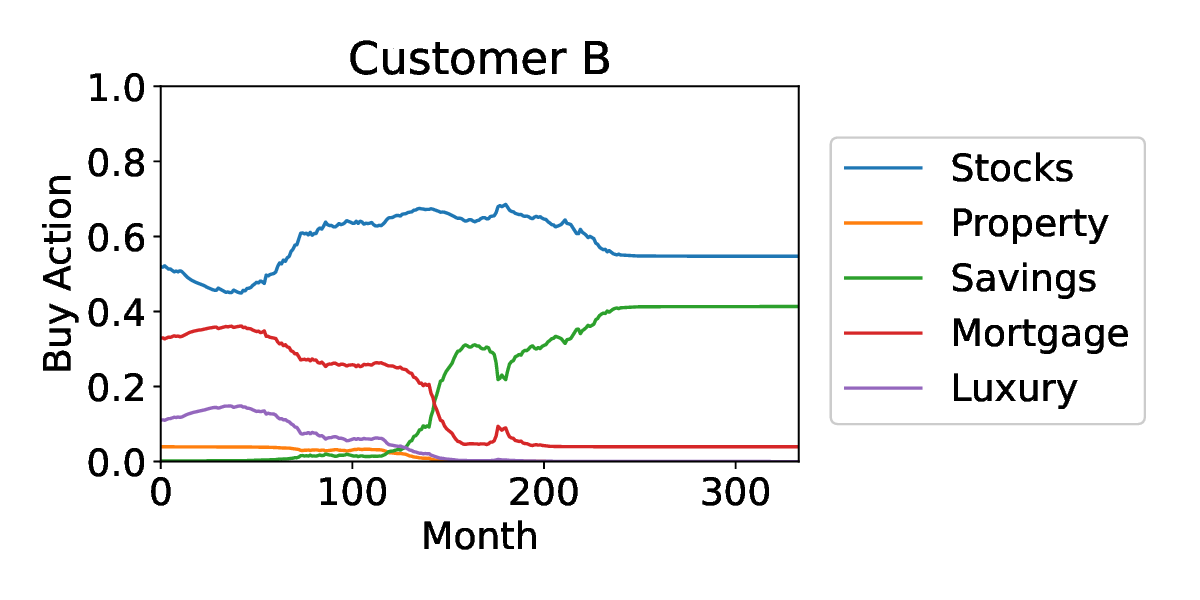}
    }\\
    \subfigure{
        \includegraphics[width=.48\linewidth]{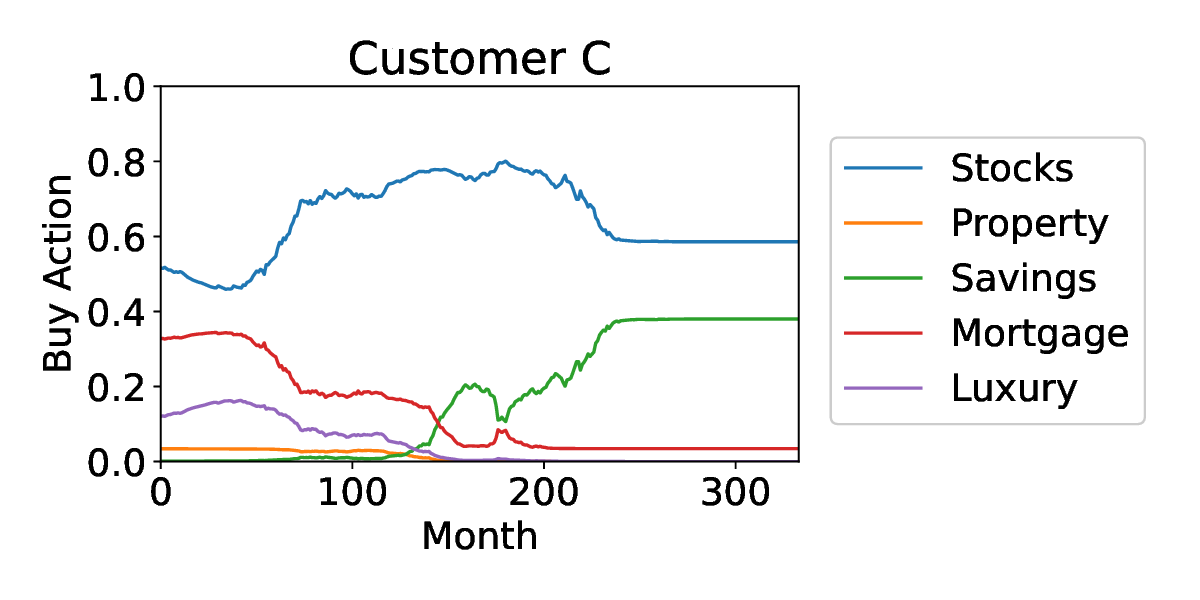}
    }\hfill
    \subfigure{
        \includegraphics[width=.48\linewidth]{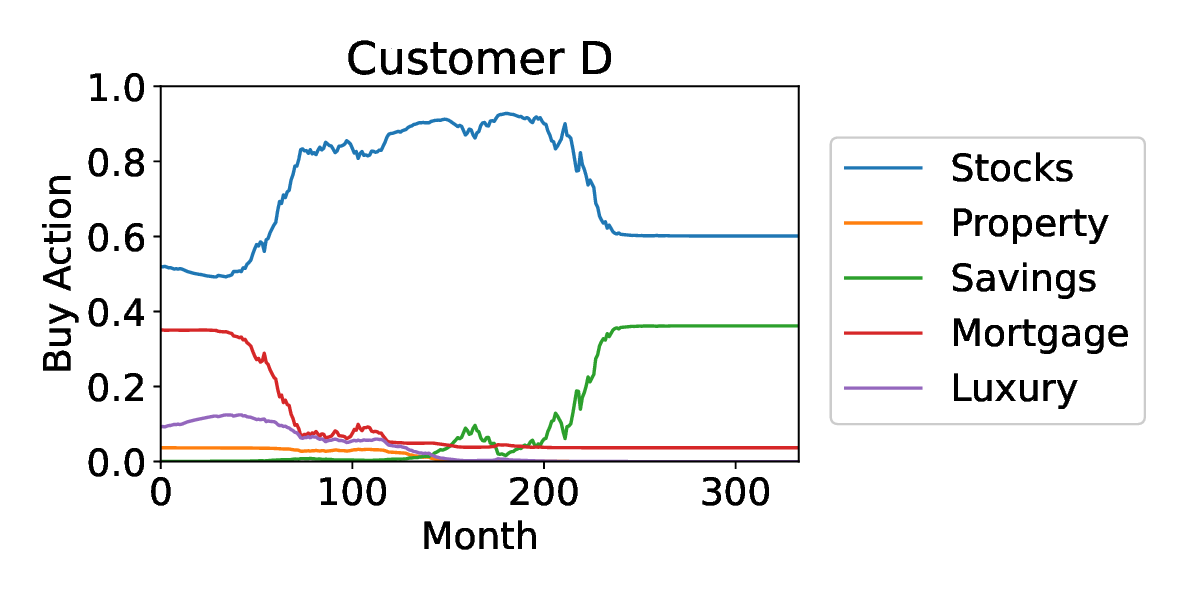}
    }
    \caption{Investment advice from four personal investment agents for four different customer personalities; they are the combined actions of the prototypical agents according to the orchestration agent. Each plot shows the investment advice in time for a single customer, named ``Customer A'' through ``Customer D'' in accordance with the labels in Figure~\ref{fig:customers}. 
    }
    \label{fig:combined_actions}
\end{figure}

These four customers' personality profiles, and consequently the orchestration agents' actions, were constant in time. Customers' personality profiles may naturally vary in time, causing directional changes in their behavioral trajectories, which alter the orchestration agent's action distribution. This affects investment strategies in real time which, for the sake of simplicity, we do not illustrate here. The investment strategies for the four customers are shown in Figure~\ref{fig:combined_actions}.  
Although these strategies might appear similar, there are significant differences: Customer~A never invested more than 60\% of their monthly allocation in stocks, while Customer~D invested up to 90\% in stocks, and Customer~A was the only one to invest significantly in property. This is due to Customer A having the highest relative degree of conscientiousness, i.e., they preferred a reduced risk. In contrast, Customer D had the highest risk in their portfolio by investing the least in property and mortgage repayments and the most in stocks, due to their low score in neuroticism which increases their appetite for risk. When comparing Customers B and C, Customer B invested more in savings accounts and less in stocks in the period between 150 and 250 months. This is due to their differences in agreeableness and neuroticism, where customer B scored higher in neuroticism and lower in agreeableness. In Figure~\ref{fig:prototypical_actions}, the prototypical agents associated with neuroticism and agreeableness are the only two to invest in savings, and the neuroticism agent started investing in savings much earlier and with higher percentages. Despite the nuanced differences in investment approaches, the general advice for all customers was similar: first pay down mortgages to reduce debt repayments, then accept higher risk with higher returns from stocks and benefit from compound growth, and finally toward retirement age reduce risk through savings accounts. This is consistent with conventional financial advice: younger people with more disposable income may accept more risk for higher returns. Very interestingly, this was not explicit in the objective function which had no elements of risk, while the effect of compound growth was evident only in increased final returns. 

The final financial returns for these four customers were very similar: after 30 years of investing 10~000 NOK per month, they all had portfolio values ranging between 21 and 24 million NOK. The total amount invested was 3.6 million NOK and the theoretical maximum return was 27.7 million NOK, achieved when investing purely in stocks. Our aim was rather to optimize customer satisfaction in their portfolio while still achieving high returns. We found that satisfaction indices between customers had greater variation than their financial returns, which we attribute to differences in the absolute values of their personality vectors. Therefore, satisfaction indices cannot be directly compared between customers, but they can be compared between different advisors for the same customer. We make such a comparison in Table~\ref{tab:perf_metrics} between the results from the orchestration agents and those of a linear combination of the prototypical agents. This linear combination is the inner product of the personality vector and the action vectors of the prototypical agents, scaled such that the resulting actions add up to one; the actions of the prototypical agents were weighted according to customers' personality vectors. The orchestration agent never performs worse than a linear combination of the prototypical agents; although it typically achieves only slightly better financial returns, it can significantly improve the satisfaction index. This was not the case when using feed-forward networks to process the customer spending input, which returned inconsistent results across multiple training runs and frequently performed worse than the simple linear combination. This is consistent with findings from \cite{Tovanich21} that spending patterns in time hold salient information not evident in non-temporal data.
\begin{table}[ht]
    \centering
    \caption{Performance metrics comparing the orchestration agent to a simple linear combination of the prototypical agents. We list the resulting portfolio values and satisfaction scores for both these scenarios after investing 10~000 NOK per month for 30 years according to the strategies shown in Figure~\ref{fig:combined_actions}.}
    \begin{tabular}{cC{2cm}C{1.6cm}C{2cm}C{1.6cm}}
         \toprule
         & \multicolumn{2}{c}{Orchestration Agent} & \multicolumn{2}{c}{Linear Combination} \\
         \cmidrule(r){2-3} \cmidrule(l){4-5}
         Customer & Portfolio Value Mill. NOK & Satisfaction Index & Portfolio Value Mill. NOK & Satisfaction Index  \\
         \midrule
         A & 20.9 & 3.1 & 20.9 & 3.0 \\
         B & 22.0 & 12.9 & 21.7 & 12.7 \\
         C & 22.7 & 19.0 & 22.4 & 17.8 \\
         D & 23.8 & 19.5 & 22.6 & 14.5 \\
         \bottomrule
    \end{tabular}
    \label{tab:perf_metrics}
\end{table}

\begin{figure}[!ht]
    \centering
    \subfigure{ 
        \includegraphics[width=.48\linewidth]{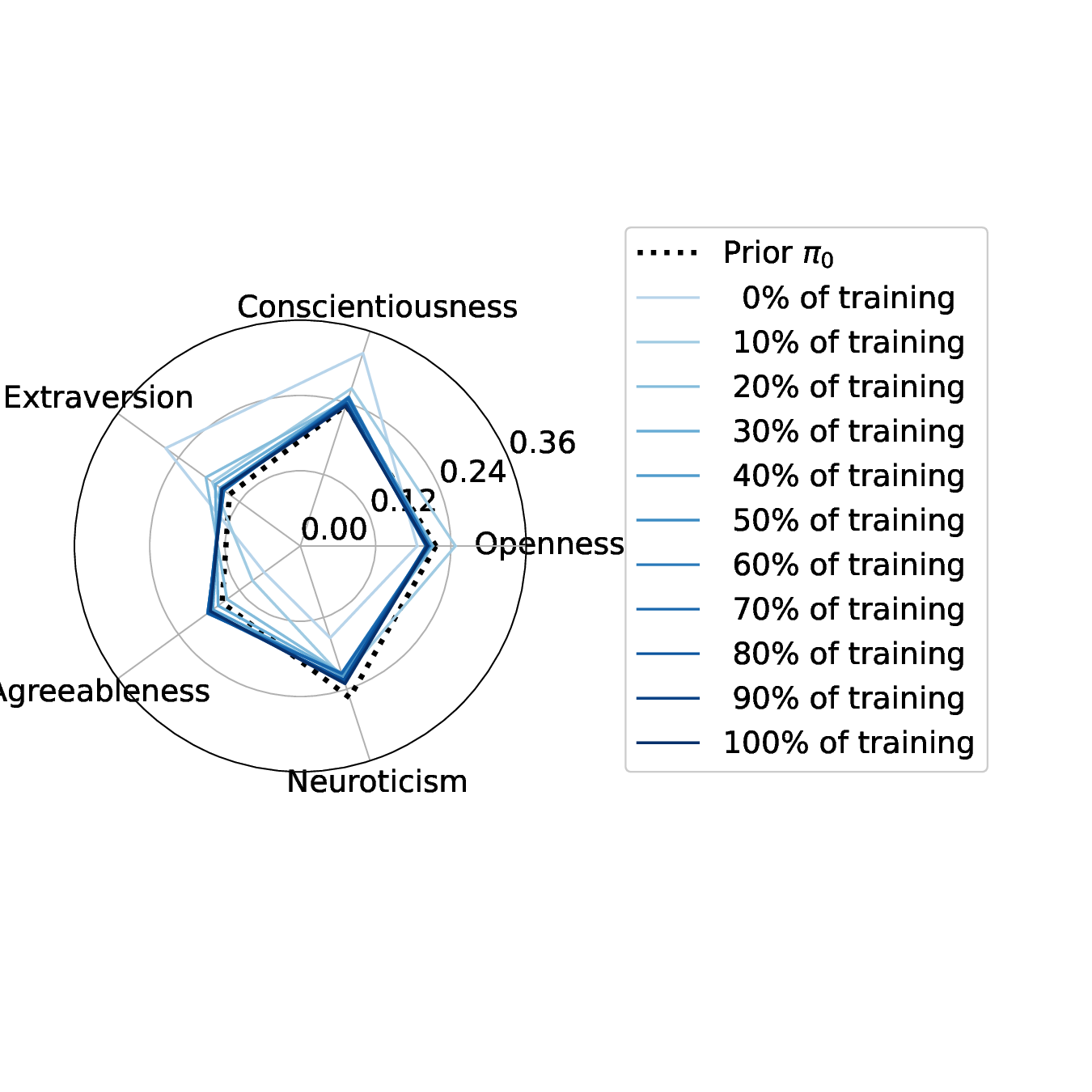}
    }\hfill
    \subfigure{
        \includegraphics[width=.48\linewidth]{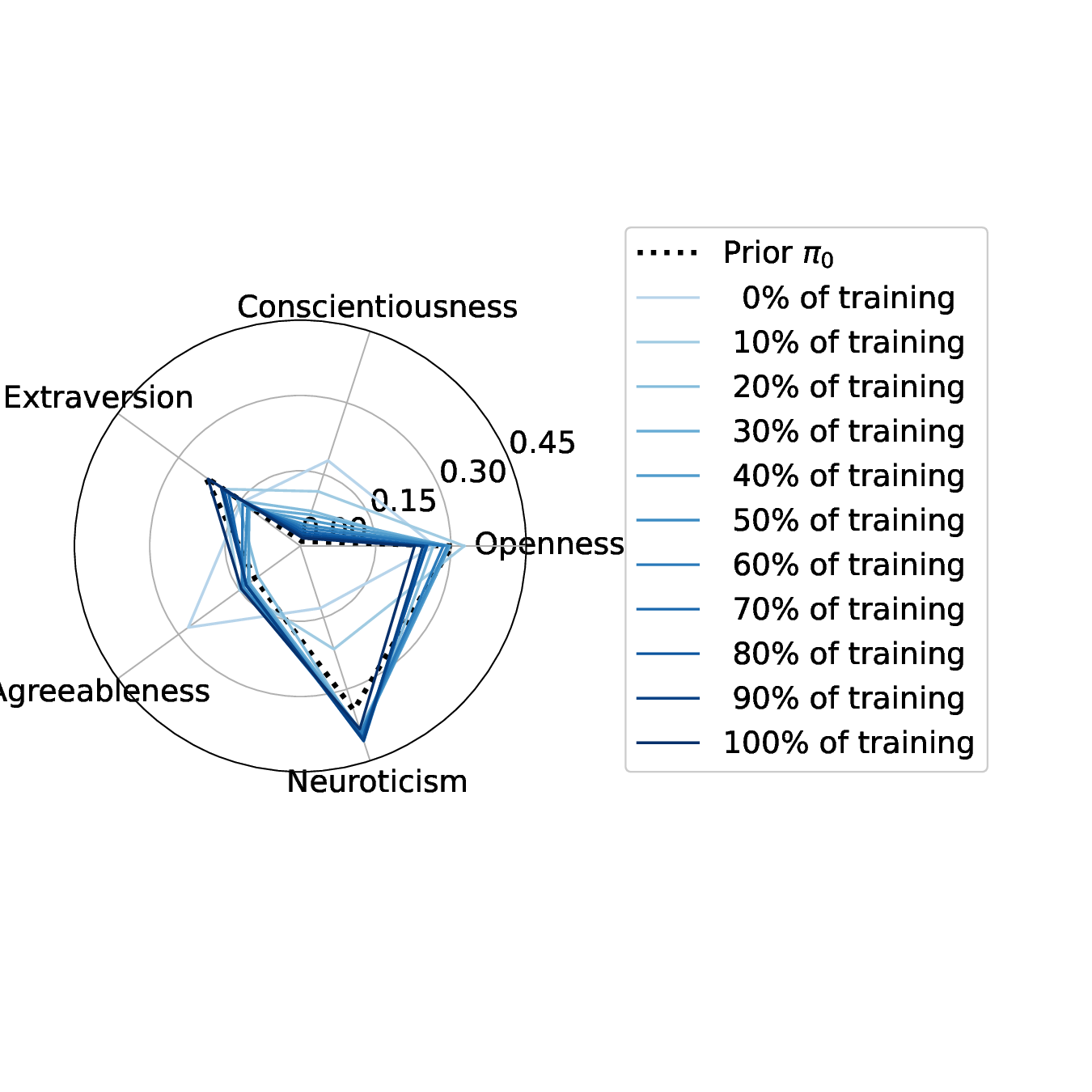}
    }\\
    \subfigure{
        \includegraphics[width=.48\linewidth]{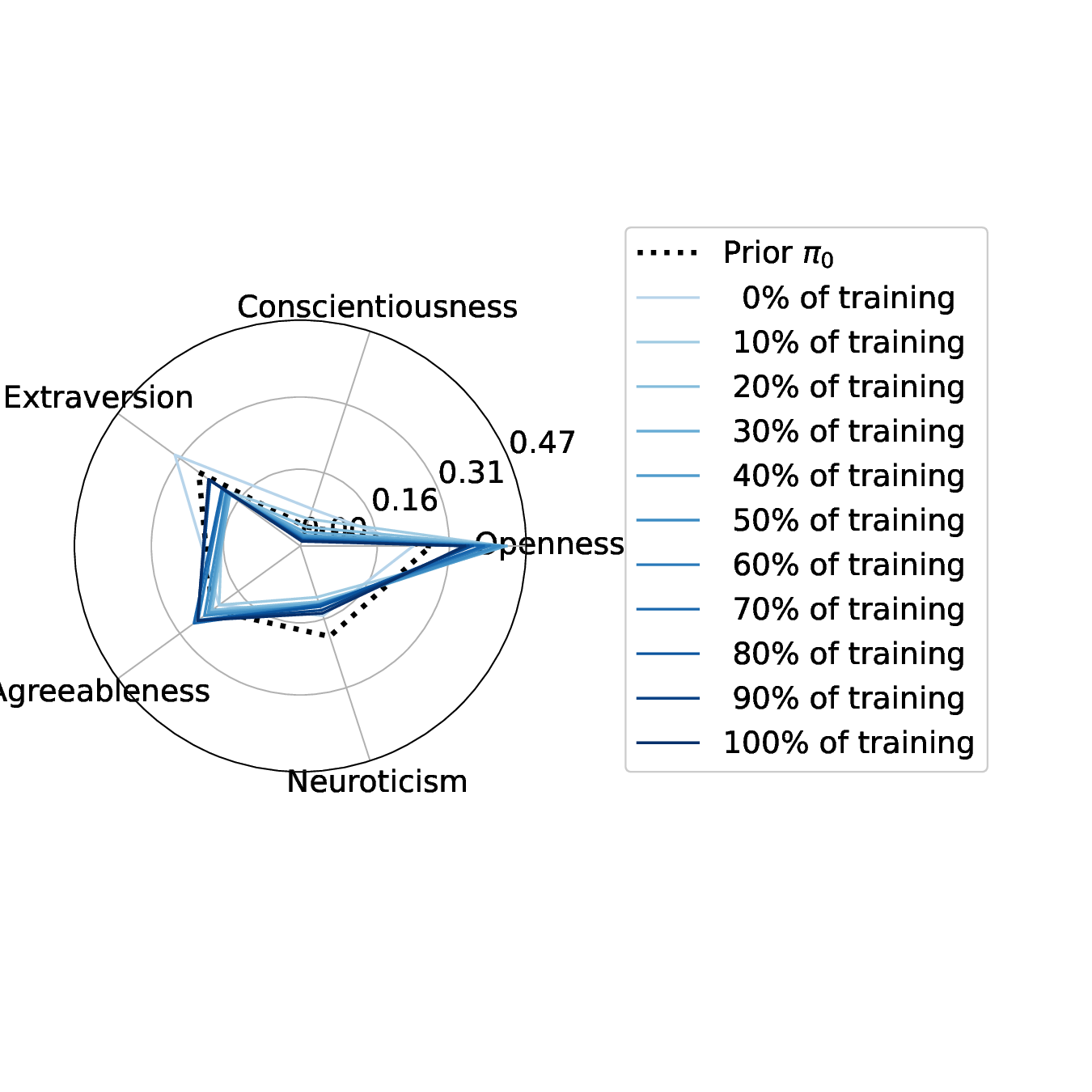}
    }\hfill
    \subfigure{
        \includegraphics[width=.48\linewidth]{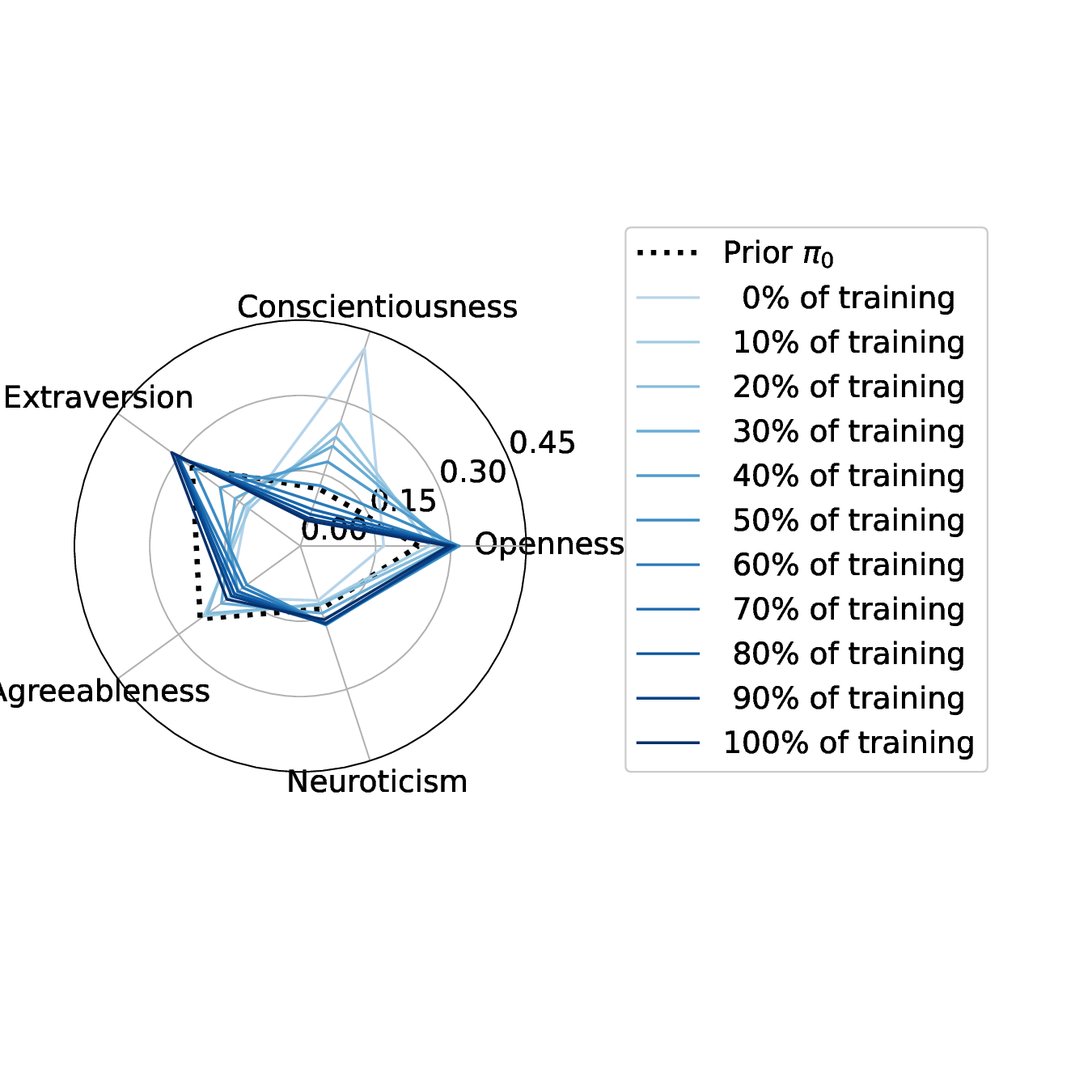}
    }
    \caption{Training convergence of the orchestration agents for four different customer personality profiles. Each successively darker blue line represents the orchestration action distribution after an increasing number of training runs. As training progresses, the successively darker blue lines converge towards the learned action distribution. The black dotted line represents the regularization prior $\pi_{0,i}$. The figures show how randomly initialized policies converge towards their specified priors and settle in a local optima in close proximity.}
    \label{fig:training_convergence}
\end{figure}

We regularized the orchestration agents to act according to a specified prior with the same action distribution as the linear combination scenario. However, through stochastic gradient descent, they optimized the satisfaction index in that region of the action space. In Figure~\ref{fig:training_convergence} we illustrate the policy convergence towards local optima of each of the four orchestration agents. 
The policies were randomly initialized, but they quickly converged to local optima in close proximity of the regularization priors in the action space. The learned strategies are thus interpretable.

\clearpage
\section{Conclusions}
Machine learning is essential for personalizing financial services. Its acceptance is contingent on understanding the underlying models, which makes model explainability and interpretability imperative. Our reinforcement learning model blends investment advice that is aligned with different personality traits. Its interpretation follows from the global intrinsic affinities of the learned policies, i.e., affinities that are independent of the current state. 
These policies not only result in a good profit, but similar profits are achieved across different personality profiles despite their distinct strategies. For instance, they avoid risk for highly conscientious individuals, while pursuing novelty for individuals that are more open to new experiences.
Interestingly, they have learned the concept of risk without this being explicit in the objective function. Across all portfolios, the advice is consistent with conventional wisdom: younger investors may accept higher risk, which typically reduces with age. It remains to be seen whether this is simply a consequence of optimizing profit while balancing the intrinsic action distribution, or whether our agents have learned deeper strategies of asset management. In future work, we intend to investigate this phenomenon by extracting an explanation for our agents' decisions. It will also be interesting to extend our method to local intrinsic affinity, where the preferred policy also depends on the current state. The potential applications for our method go beyond investment advice and includes, e.g., preferred evasive actions for self driving cars, navigation systems that favor either freeways or scenic roads, or medication for the treatment of chronic diseases based on individual medical history in healthcare.

\subsection*{Declarations}
\begin{itemize}
    \item \textbf{Funding}: This study was partially funded by a grant from The Norwegian Research Council, project number 311465.
    \item \textbf{Competing Interests}: The authors declare no competing interests.
    \item \textbf{Ethics approval}: Not applicable.
    \item \textbf{Consent to participate}: Personal data were anonymized and processing was done on the basis of consent in compliance with the European General Data Protection Regulation (GDPR).
\end{itemize}

\bibliographystyle{unsrt}
\bibliography{references}
\end{document}